\documentclass[10pt,journal,cspaper,compsoc]{IEEEtran}

\usepackage{amsmath}
\usepackage{graphicx}
\usepackage{subfig}
\usepackage{cite}

\usepackage{tablefootnote}

\ifCLASSOPTIONcompsoc
\else
\fi
\ifCLASSINFOpdf
\else
\fi
\begin{document}
%
\title{A Generic Method for Automatic Ground Truth Generation of Camera-captured Documents}
%
%
%
%

\author{Sheraz~Ahmed, Muhammad~Imran~Malik, Muhammad~Zeshan~Afzal, Koichi~Kise, Masakazu~Iwamura,  Andreas~Dengel, Marcus~Liwicki´
\IEEEcompsocitemizethanks{\IEEEcompsocthanksitem Sheraz~Ahmed, Muhammad~Imran~Malik, Muhammad~Zeshan~Afzal,  Andreas Dengel, and Marcus~Liwicki are with German Research Center for Artificial Intelligence (DFKI GmbH), Germany and Technische Universit{\"a}t Kaiserslautern, Germany.
\protect\\
E-mail: firstname.lastname@dfki.de
\IEEEcompsocthanksitem Koichi~Kise and Masakazu~Iwamura are with Osaka Prefecture University, Japan}
\thanks{}}

\IEEEcompsoctitleabstractindextext{%
\begin{abstract}

The contribution of this paper is fourfold. 
The first contribution is a novel, generic method for automatic ground truth generation of camera-captured document images (books, magazines, articles, invoices, etc.).
It enables us to build large-scale (i.e., millions of images) labeled camera-captured/scanned documents datasets, without any human intervention.
The method is generic, language independent and can be used for generation of labeled documents datasets (both scanned and camera-captured) in any cursive and non-cursive language, e.g., English, Russian, Arabic, Urdu, etc.
To assess the effectiveness of the presented method, two different datasets in English and Russian are generated using the presented method.
Evaluation of samples from the two datasets shows that $99.98\%$ of the images were correctly labeled.
The second contribution is a large dataset (called C$^3$W\textit{i}) of camera-captured characters and words images, comprising $1$ million word images ($10$ million character images), captured in a real camera-based acquisition.
This dataset can be used for training as well as testing of character recognition systems on camera-captured documents.
The third contribution is a novel method for the recognition of camera-captured document images. The proposed method is based on Long Short-Term Memory and outperforms the state-of-the-art methods for camera based OCRs.
As a fourth contribution, various benchmark tests are performed to uncover the behavior of commercial (ABBYY), open source (Tesseract), and the presented camera-based OCR using the presented C$^3$W\textit{i} dataset.
Evaluation results reveal that the existing OCRs, which already get very high accuracies on scanned documents, have limited performance on camera-captured document images; where ABBYY has an accuracy of 75\%, Tesseract an accuracy of 50.22\%, while the presented character recognition system has an accuracy of 95.10\%.

\end{abstract}

\begin{keywords}
Camera-captured document, Automatic ground truth generation, Dataset, Document image degradation, Document image retrieval, LLAH, OCR.

\end{keywords}}

\maketitle

\IEEEdisplaynotcompsoctitleabstractindextext

%
\IEEEpeerreviewmaketitle

\section{Introduction}

Text recognition is an important part in the analysis of camera-captured documents as there are plenty of services which can be provided based on the recognized text. For example, if text is recognized, one can provide real time translation and information retrieval.
Many Optical Character Recognition systems (OCRs) available in the market~\cite{ABBYY,Omnipage,Tesseract1,OCROpus} are designed and trained to deal with the distortions and challenges specific to scanned document images.
 
However, camera-captured document distortions (e.g., blur, perspective distortion, occlusion) are different from those of scanned documents.
To enable the current OCRs (developed originally for scanned documents) for camera-captured documents, it is required to train them with data containing distortions available in camera-captured documents.

The main problem in building camera based OCRs is the lack of publicly available dataset that can be used for training and testing of character recognition systems for camera-captured documents~\cite{PhotoOCR}. 
One possible solution could be to use different degradation models to build up a large-scale dataset using synthetic data~\cite{generative:camera,generative:camera1}.
However, researchers are still of different opinions about either degradation models are true representative of real world data or not.
Another possibility could be to generate this dataset by manually extracting words and/or characters from real camera-captured documents and labeling them. 
However, the manual labeling of each word and/or character in captured images is impractical for being very laborious and costly. 
Hence, there is a strong need of automatic methods capable of generating datasets from real camera-captured text images.

Some methods are available for automatic labeling/ ground truth generation of scanned document images~\cite{joost:newspaper,joost:OCR,Kanungo:chargt,Kanungo98anautomatic,LAMP:GT}. These methods mostly rely on aliging scanned documents with the existing digital versions. However, the existing methods for ground truth generation of scanned documents cannot be applied to camera-captured documents, as they assume that the whole document is contained in the scanned image. In addition, these methods are not capable of dealing with problems mostly specific to camera-captured images (blur, perspective distortion, occlusion).

This paper presents a generic method for automatic labeling/ground-truth generation of camera-captured text document images using a document image retrieval system.
The proposed method is automatic and does not require any human intervention in extraction/localization of words and/or characters and their labeling/ground truth generation. 
A Locally Likely Arrangement Hashing (LLAH) based document retrieval system is used to retrieve and align the electronic version of the document with the captured document image.
LLAH can retrieve the same document even if only a part of document is contained in the camera-captured image~\cite{LLAH:takeda}.

The presented method is generic and script independent. This means that it can be used to build documents (both camera-captured and scanned) datasets for different languages, e.g., English, Russian, Japanese, Arabic, Urdu, Indic scripts, etc.
All we need is PDF (electronic version) of documents and their camera-captured/scanned image. 
To test the method, we have successfully generated two datasets of camera-captured documents in English and Russian, with an accuracy of 99.98\%

In addition to a ground truth generation method, we introduce a novel, large, word and character level dataset consisting of one million words and ten million character images extracted from camera-captured text documents. These images contain real distortions specific to camera-captured images (e.g., blur, perspective distortion, varying lighting). 
The dataset is generated automatically using the presented automatic labeling method. 
We refer this dataset as Camera-Captured Characters and Words images (C$^{3}$W\textit{i}) dataset.

To show the impact of the presented dataset, we presented and evaluated a Long Short Term Memory (LSTM) based character recognition system that is capable of dealing with the camera based distortion and outperforms both commercial (ABBYY) as well as open source (Tesseract) OCRs by achieving a recognition accuracy of more than 97\%.
The presented character recognition system is not specific to only camera-captured images but also performs reasonably well on scanned document images by using the same model trained for camera-captured document images. 

Furthermore, we have also evaluated both commercial as well as open source OCR systems on our novel dataset. The aim of this evaluation is to uncover the behavior of these OCRs on real camera-captured document images. The evaluation results show that there is a lot of room for improvements in OCR for camera-captured document images in presence of quality related distortion (blur, varying lighting conditions, etc.).

\section{Related Work}\label{sec:relatedwork}
This section provides an overview of different available datasets and summarizes different approaches for automatic ground truth generation.
First, Section~\ref{sec:existingdataset} provides an overview of different datasets available for camera-captured documents and natural scene images.
Second, Section~\ref{sec:gtmethods} provides details about different degradation models for scanned and camera-captured images.
In addition, it also provides review of the various existing approaches for automatic ground truth generation.

\subsection{Existing Datasets}\label{sec:existingdataset}
To the best of authors' knowledge, currently there is no 'publicly' available dataset for camera-captured text document images (like books, magazines, article, newspaper) which can be used for training of character recognition systems on camera-captured documents.

Bukhari et al.~\cite{bukhari11} has introduced a dataset of camera-captured document images. This dataset consist of $100$ pages with the text line information. In addition, ground truth text for each page is also provided. It is primarily developed for text line extraction and dewarping. It cannot be used for training of character recognition systems because there is no character, word, or line level text ground truth information available.
Kumar et al.~\cite{JayantCBDAR13} has proposed a dataset containing $175$ images of $25$ documents taken with different camera settings and focus. This dataset can be used for assessing the quality of images, e.g., sharpness score. However, it cannot be used for training of OCRs on camera-captured documents, as there is no character, word, or line level text ground truth information available.
Bissacco et al.~\cite{PhotoOCR} has used a dataset of manually labeled documents which were submitted to Google for queries. However, the dataset is not publicly available, and therefore cannot be used for improving other systems.  

Recently, a camera-captured document OCR competition is organized in ICDAR 2015 with the focus on evaluation of text recognition from images captured by mobile phones~\cite{icdar15:smartdoc}. This dataset contains single column 12100 camera-captured document images in English with manually transcribed OCR ground truth (raw text) for complete pages. Similar to Bukhari et al.~\cite{bukhari11}, it cannot be used to train OCRs because there is no character, word, or line level text ground truth information available.

\begin{figure}[b]
  \centering  
	\subfloat[][]{\includegraphics[width=0.55\columnwidth]{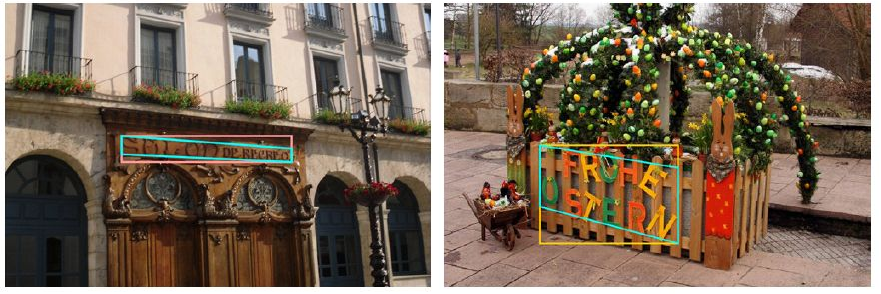}}
	\hspace{0.2cm}
	\subfloat[][]{\includegraphics[width=0.35\columnwidth]{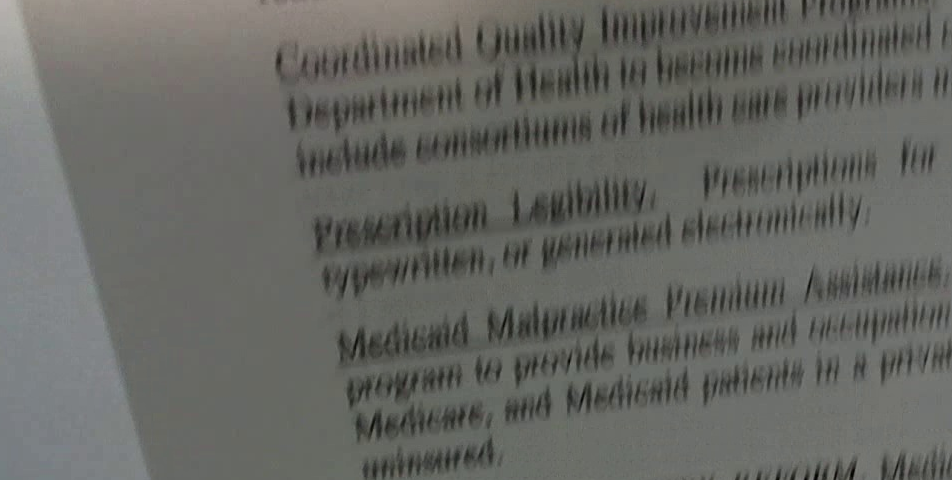}}
\caption{Samples of text in (a) Natural scene image and (b) Camera-captured document image}
 \label{fig:samples:nature-camera}
\end{figure}

\begin{figure*}[t]
  \centering  
	\subfloat[][]{\includegraphics[width=0.22\textwidth,height=2.3cm]{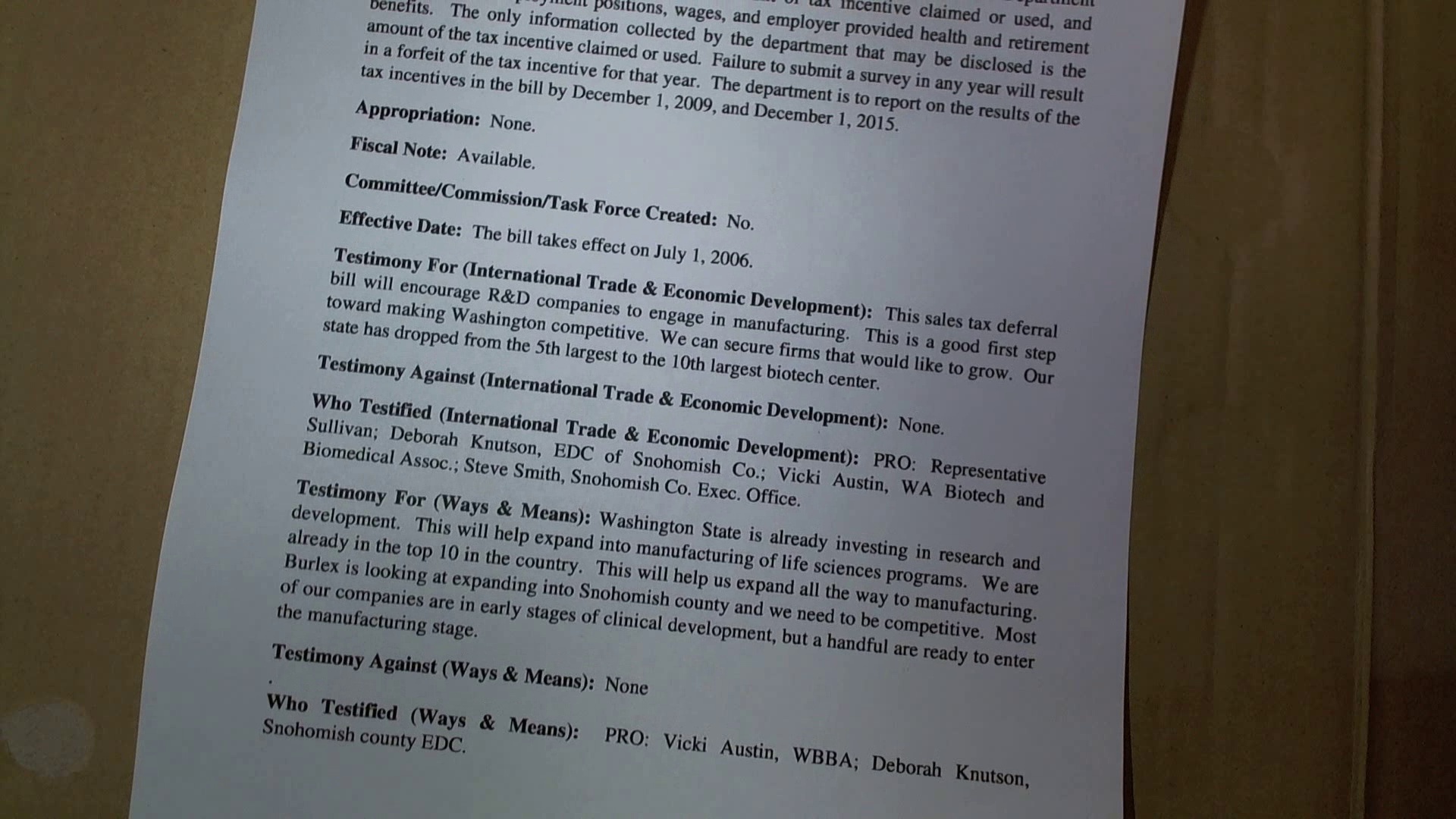}}
	\hspace{0.5cm}
	\subfloat[][]{\includegraphics[width=0.17\textwidth,height=2.3cm]{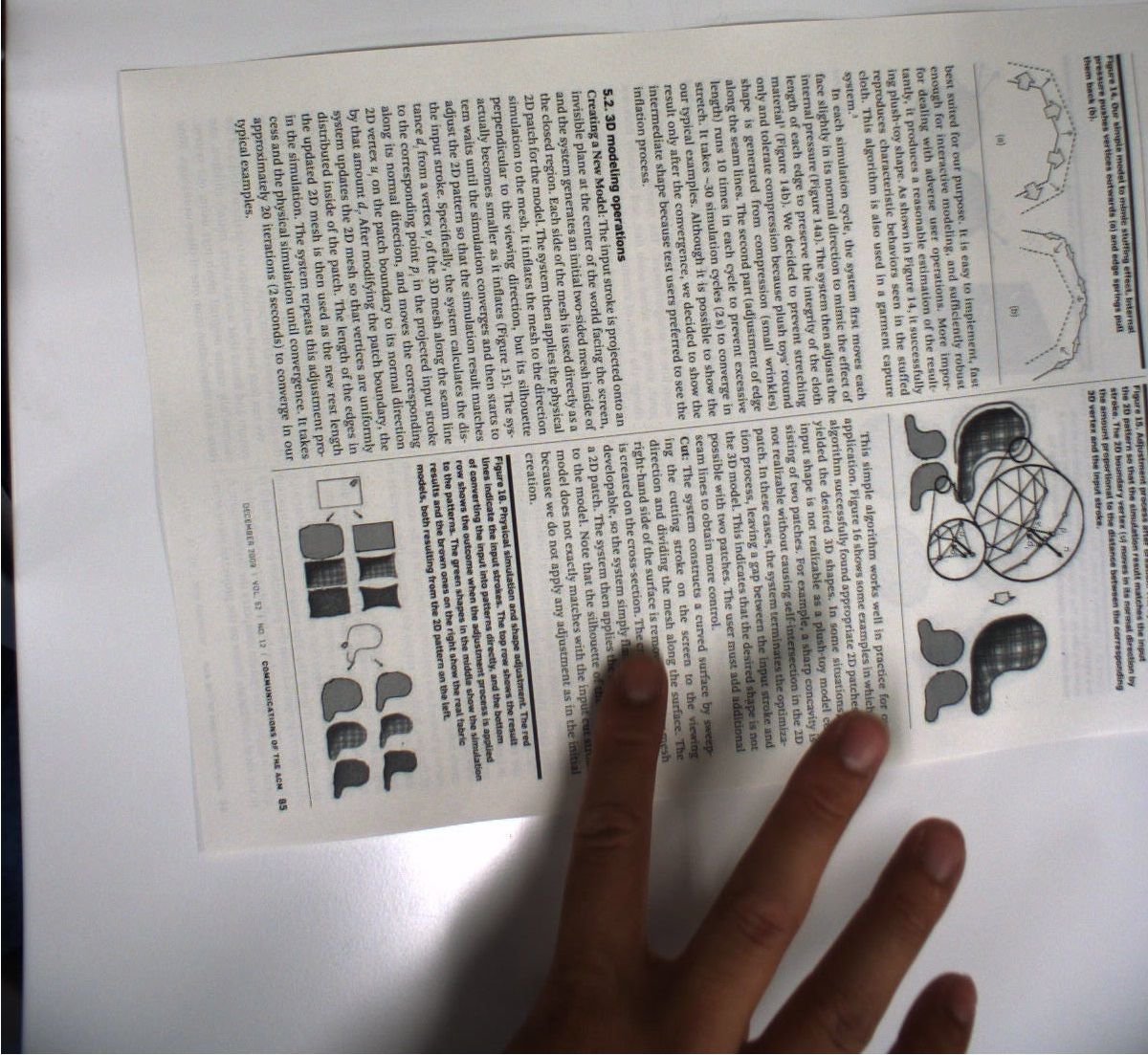}}
	\hspace{0.5cm}
	\subfloat[][]{\includegraphics[width=0.22\textwidth,height=2.3cm]{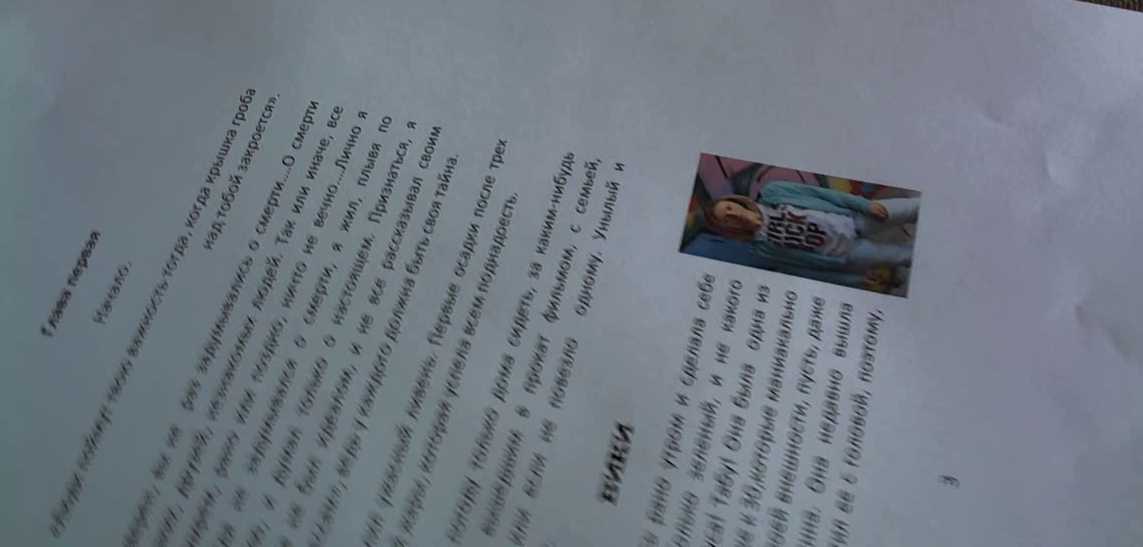}}
	\hspace{0.5cm}
	\subfloat[][]{\includegraphics[width=0.22\textwidth,height=2.3cm]{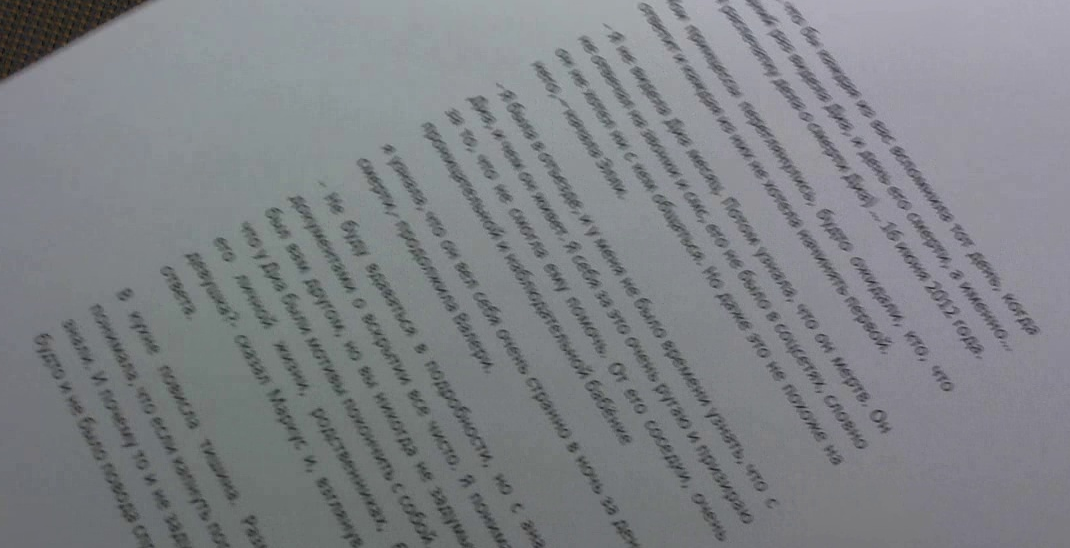}}
\caption{Samples of camera-captured documents in English (a,b) and Russian (c,d)}
 \label{fig:samples}
\end{figure*}

In the last few years, text recognition in natural scene images has gained a lot of attention of researchers. In this context different datasets and systems are developed. 
The major datasets available are the ones from series of ICDAR Robust Reading Competitions~\cite{ICDAR_RRC2003_IJDAR2005,ICDAR2011_Robust_Reading_Competition_challenge2,ICDAR2013_RRC,ICDAR2015_RRC}.
The focus is to enable text recognition in natural scene images, where text is present as either embossed on objects, merged in the background, or is available in arbitrary forms. Figure~\ref{fig:samples:nature-camera} (a) shows natural scene images with text. 
Similarly, de Campos et al.~\cite{char64k} proposed a dataset consisting up of symbols used in both English and Kannada. It contains characters from natural images, hand drawn characters on tablet PC, and synthesized characters from computer fonts. 
Netzer et al.~\cite{NIPS11} introduced a dataset consisting of digits extracted from natural images. The numbers are taken from house numbers in the Google Street View images, and therefore the dataset is known as the Street View House Numbers (SVHN) dataset. 
However, it only contains digits from natural scene images.
Similarly, Nagy et al.~\cite{NEOCR12a} proposed a Natural Environment OCR (NEOCR) dataset with a collection of real world images depicting text in different natural variations. Word level ground truth is marked inside the natural images. 
All of the above-mentioned datasets are developed to deal with text recognition problem in natural images. 
However, our focus is on documents like books, newspapers, magazines, etc., captured using camera, with different camera related distortions e.g., blur, perspective distortion, and occlusion. (Figure~\ref{fig:samples:nature-camera} (a) shows example images from natural scenes with text while Figure~\ref{fig:samples} shows example camera-captured document images). None of the above mentioned datasets contains any samples from the documents similar to those in Figure~\ref{fig:samples:nature-camera} (b) and Figure~\ref{fig:samples}). Therefore, these datasets cannot be used for training of OCRs with the intention to make them working on camera-captured document images.  

\subsection{Ground Truth Generation Methods}\label{sec:gtmethods}
One popular method for automatic ground truth generation is to use different image degradation models~\cite{degradationModels:Baird,Baird00thestate}.
An advantage of degradation models is that everything remains electronic, so we do not need to print and scan documents.
Degradation models are applied to word or characters to generate images with different possible distortions.
Zi \cite{LAMP:GT} used degradation models to synthetic data in different languages, for building datasets, which can be used for training and testing of OCR. 
Furthermore, some image degradation models have also been proposed for camera-captured documents.
Tsuji et al.~\cite{generative:camera} has proposed a degradation model for low-resolution camera-captured character recognition.
The distribution of the degradation parameters is estimated from real images and then applied to build synthetic data.
Similarly, Ishida et al.~\cite{generative:camera1} proposed a degradation model of uneven lighting which is used for generative learning.
The main problem with degradation models is that they are designed to add limited distortions estimated from distorted image.
Thus, it is still debatable that either these models are true representative of real data or not.

In addition to the use of different degradation models, another possibility is to use alignment-based methods where real images are aligned with electronic version to generate ground truth. 
Kanungo \& Haralick \cite{Kanungo:chargt,Kanungo98anautomatic} proposed an approach for character level automatic ground truth generation from scanned documents. Documents are created, printed, photocopied, and scanned.
Geometric transformation is computed between scanned and ground truth images.
Finally, transformation parameters are used to extract the ground truth information for each character.
Kim \& Kanungo~\cite{Kim} further improved the approach presented by Kanungo \& Haralick\cite{Kanungo:chargt,Kanungo98anautomatic} by using attributed branch-and-bound algorithm for establishing correspondence between the data points of scanned and ground truth images. After establishing the correspondence, ground truth for the scanned image is extracted by transforming the ground truth of the original image.

Similarly, Beusekom et al.~\cite{joost:OCR} proposed automatic ground truth generation for OCR using robust branch and bound search (RAST)~\cite{RASTBreuel}.
First, global alignment is estimated between the scanned and ground	 truth images.
Then, local alignment is used to adapt the transformation parameters by aligning clusters of nearby connected components.
Strecker et al.~\cite{joost:newspaper} proposed an approach for ground truth generation of newspaper documents.
It is based on synthetic data generated using an automatic layout generation system. The data are printed, degenerated, and scanned.
 Again, RAST is used to compute the transformation to align the ground truth to the scanned image.
The focus of this approach is to create ground truth information for layout analysis.

Note that in the case of scanned documents, complete document image is available, and therefore, transformation between ground truth and scanned image can be computed using alignment techniques mentioned in~\cite{joost:newspaper,joost:OCR,Kanungo:chargt,Kanungo98anautomatic}.
However, camera-captured documents usually contain mere parts of documents along with other, potentially unnecessary, objects in the background. Figure~\ref{fig:samples} shows some samples of real camera-captured document images.
Here, application of the existing ground truth generation methods is not possible due to partial capture and perspective distortions. Note that for scanned document images, mere scale, translation, and rotation (similarity transformation) is enough which is contrary to camera-captured document images.

Recently, Chazalon et al.~\cite{icdar15:semiauto} proposed a semi-automatic method for segmentation of camera/mobile captured document image based on color markers detection.
Up to the best of authors' knowledge, there is no method available for automatic ground truth generation for camera-captured document images. This makes the contribution of this paper substantial for the document analysis community.

\section{Automatic Ground Truth Generation : The Proposed Approach}\label{sec:Methodology}
The first step in any automatic ground truth generation methods is to associate camera-captured images with their electronic versions.
In the existing methods for ground truth generation of scanned documents, it is required to manually associate the electronic version of document with the scanned image so that they could be aligned. 
This manual association limits the efficiency of these methods.

To overcome the manual association and to make the proposed method fully automatic, we used a document image retrieval system. 
This document image retrieval system automatically retrieves the electronic versions of the camera-captured document images. 
Therefore, to generate the ground truth, the only thing to do is to capture images of the documents.  
In the proposed method, an LLAH based document retrieval system is used for retrieving the electronic version of the camera-captured text document.
This part is referred to as document level matching, Section~\ref{sec:Methodology:ret} provides an overview of this step.

After retrieving the electronic version of a camera-captured document, the next step is to align the camera-captured document with its electronic version.
The application of existing alignment methods is not possible on camera-captured documents because of 'partial capture' and 'perspective distortion'. 
To align a camera-captured document with its electronic version, it is required to perform the following steps: 
\begin{itemize}
	\item \textbf{Estimate the regions in electronic version that correspond to camera-captured document.}
	This estimation is necessary for aligning the parts of electronic version which correspond to a camera-captured document image.
	It is performed with the help of LLAH, as it not only retrieves the electronic version of captured document, but also provides the estimate of the region/part of electronic version of the document corresponding to the captured document. Section~\ref{sec:Methodology:ret} provides details about LLAH.
	\item \textbf{Alignment of camera-captured document with its corresponding part in electronic document}.

Using the corresponding region/part estimated by LLAH, part level matching and transformations are performed to align the electronic and the captured image. 
Section~\ref{sec:Methodology:gt} provides details about part level matching.
	\item \textbf{Words alignment and ground truth extraction}
	
Finally, using the parts of image from both the camera-captured and the electronic version of a document, word level matching and transformation is performed to extract corresponding words in both images and their ground truth information from PDF.
Section~\ref{sec:Methodology:word} provides details of word level matching.
This step results into word and character images along with their ground truth information. 
	\end{itemize}

\subsection{Document Level Matching}\label{sec:Methodology:ret}

The electronic version of the captured document is required to align a camera-captured document with its electronic version.
In the proposed method, we have automated this process by using document level matching. Here, the electronic version of a captured document is automatically retrieved from the database by using an LLAH based document retrieval system.
LLAH is used to retrieve the same document from large databases with efficient memory scheme.
It has already shown the potential to extract similar documents from the database of $20$ million images with retrieval accuracy of more than $99\%$~\cite{LLAH:takeda}.
\begin{figure}[h]
  \centering  
\includegraphics[width=\columnwidth]{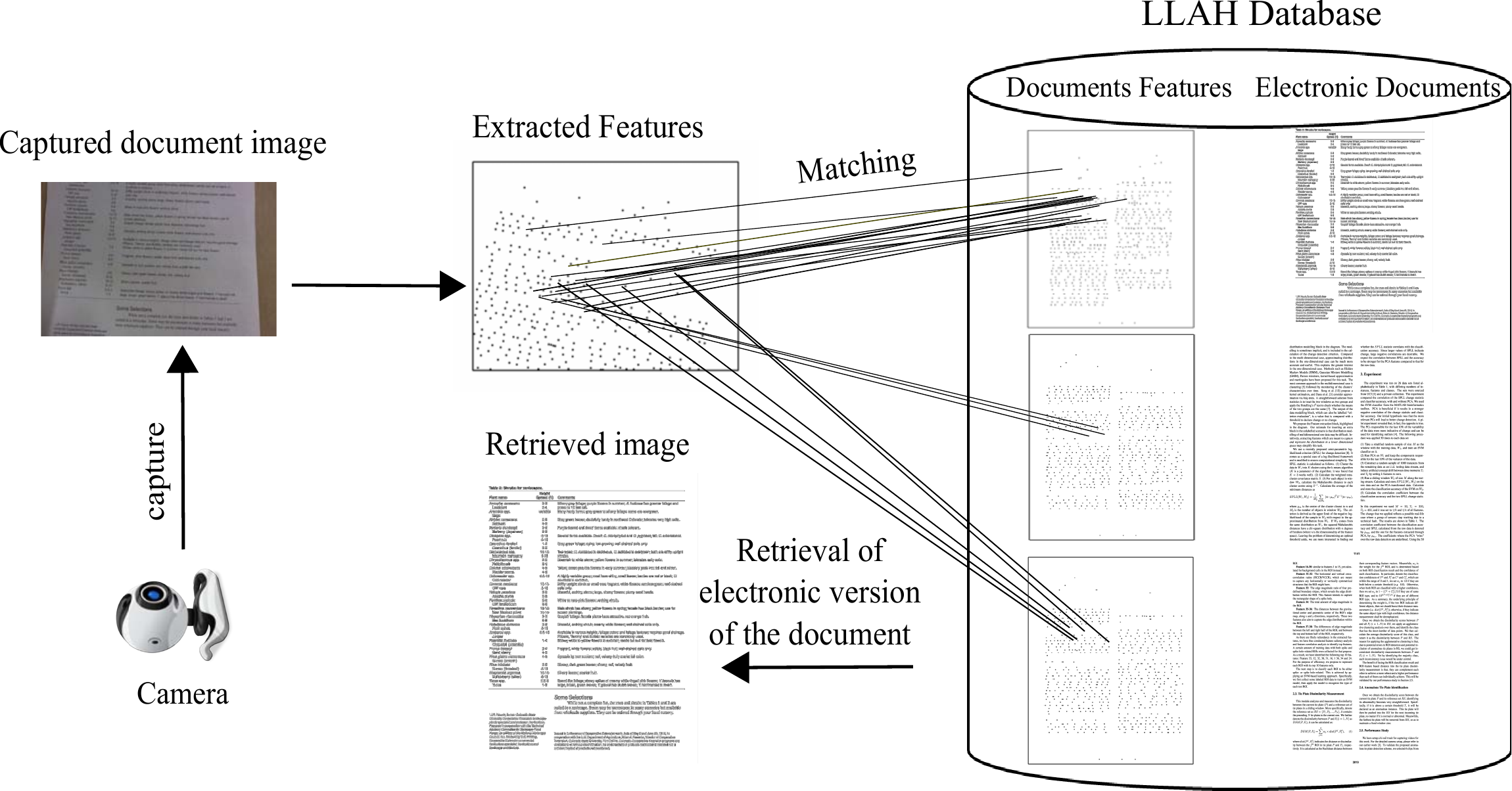}
	\caption{Document retrieval with LLAH}
 \label{fig:dlmatching}
\end{figure}

Figure~\ref{fig:dlmatching} shows the LLAH based document retrieval system.
To use document retrieval system, it is required to first build a database containing electronic version of documents. 
To build the database, document images are rendered from their corresponding PDF files at $300$ dpi. 
The documents used to build this database include, proceedings, books, and magazines.

Here we are summarizing LLAH for completeness; details can be found in~\cite{LLAH:takeda}.
The LLAH extracts local features from camera-captured documents. These features are based on the ratio of the areas of two adjoined triangles made by four coplanar points.
First, Gaussian blurring is applied on the camera-captured image which is then converted into feature points (centroid of each connected component). The feature vector is calculated at each feature point by finding its $'n'$ nearest points. Then $'m'$ $ (m \leq n)$ points are chosen from those $'n'$ points, and among these $'m'$ points, four are chosen at a time to calculate the affine invariance. This process is repeated for all combinations and 4 from m are chosen. Hence, each feature point will result in $\left(\begin{array}{c}n\\ m\end{array}\right)$  descriptors and each descriptor is of $\left(\begin{array}{c}m\\ 4\end{array}\right)$ dimensions. 
To efficiently match feature vectors, LLAH employs hashing of feature vectors. 
To obtain the hash index, discretization is performed on the descriptors. Finally, the document ID, point ID, and the discretized feature are stored in a hash table according to the hash index.
Hence, each entry in the hash table corresponds to a document with its features.

To retrieve the electronic version of a document from the database, features are extracted from the camera-captured image and compared to features in the database.
Electronic version (PDF and image) of the document, having the highest matching score, is returned as the retrieved document.

\subsection{Part level Matching}\label{sec:Methodology:gt}

Once electronic version of a camera-captured document is retrieved. 
The next step is to align the camera-captured document with its electronic version. To do so, it is required to estimate the region of electronic document image (retrieved by document retrieval system) which corresponds to the camera-captured image.
This region is computed by making a polygon around the matched points in electronic version of document~\cite{LLAH:takeda}. 
Using this corresponding region, the electronic document image is cropped so that only the corresponding part is used for further processing.
\begin{figure}[t]
  \centering  
\includegraphics[width=\columnwidth]{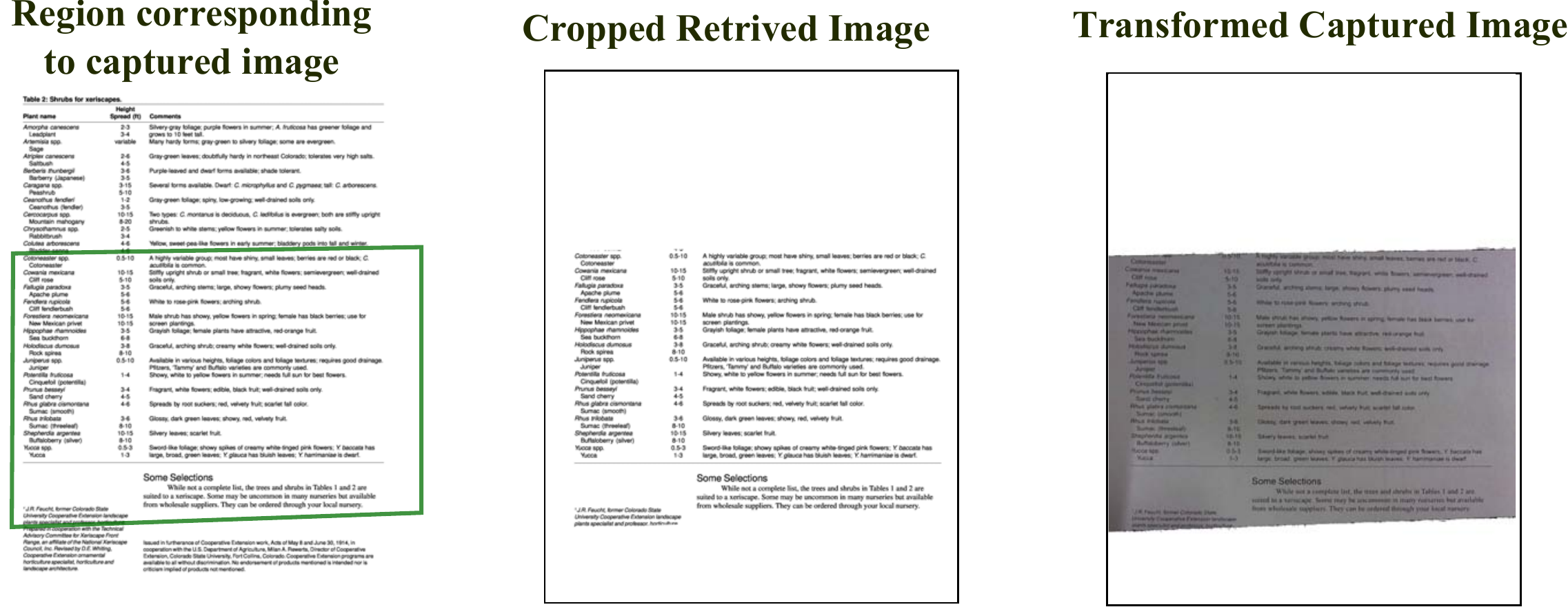}
	\caption{Estimation and alignment of document parts}
 \label{fig:plmatching}
\end{figure}

To align these regions and to extract ground truth, it is required to first map them into the same space.
As compared to scanned documents, camera-captured images contain different types of distortions and transformations (Figure~\ref{fig:samples}). Therefore, we need to find out transformation parameters which can convert the camera-captured image to the electronic image space.
The transformation parameters are computed by using the least square method on the corresponding matched points between the query and the electronic/retrieved version of document image.
The computed parameters are further refined with the Levenberg-Marquardt method~\cite{levenberg44} to reduce the re-projection error.
Using these transformation parameters, perspective transformation is applied to the captured image, which maps it to the space of the retrieved document image.

Figure~\ref{fig:plmatching} shows the cropped electronic document image and the transformed/normalized captured images (captured image after applying perspective transformation) which are further used in word level processing to extract ground truth.

\subsection{Word Level Matching and Ground Truth Extraction}\label{sec:Methodology:word}
\begin{figure}[t!]
  \centering  
	\includegraphics[width=0.7\columnwidth]{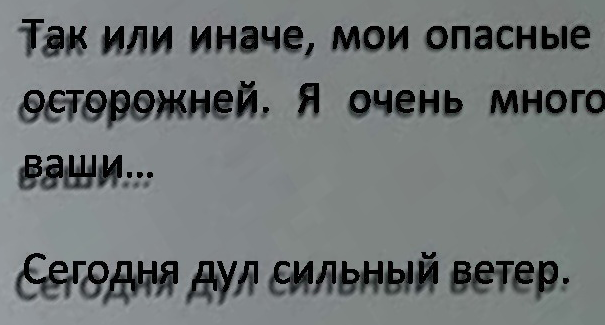}
	\caption{Overlapped electronic version and normalized camera-captured images}
 \label{fig:overlap}
\end{figure}

Figure~\ref{fig:overlap} shows the aligned camera-captured and electronic documents. 
It can be seen that only some parts of both documents (electronic and transformed captured) are perfectly aligned.
This is because; the transformation parameters provided by the LLAH are approximated parameters and are not perfect.
If these transformation parameters were directly used to extract corresponding ground truth from PDF file, it would lead to false ground truth information for the parts which are not perfectly aligned.
The word level matching is performed to avoid this error. Here, the perfectly aligned regions are located so that exactly the same and complete word is cropped from the captured and electronic images.

\begin{figure}[t]
  \centering  
\includegraphics[width=\columnwidth]{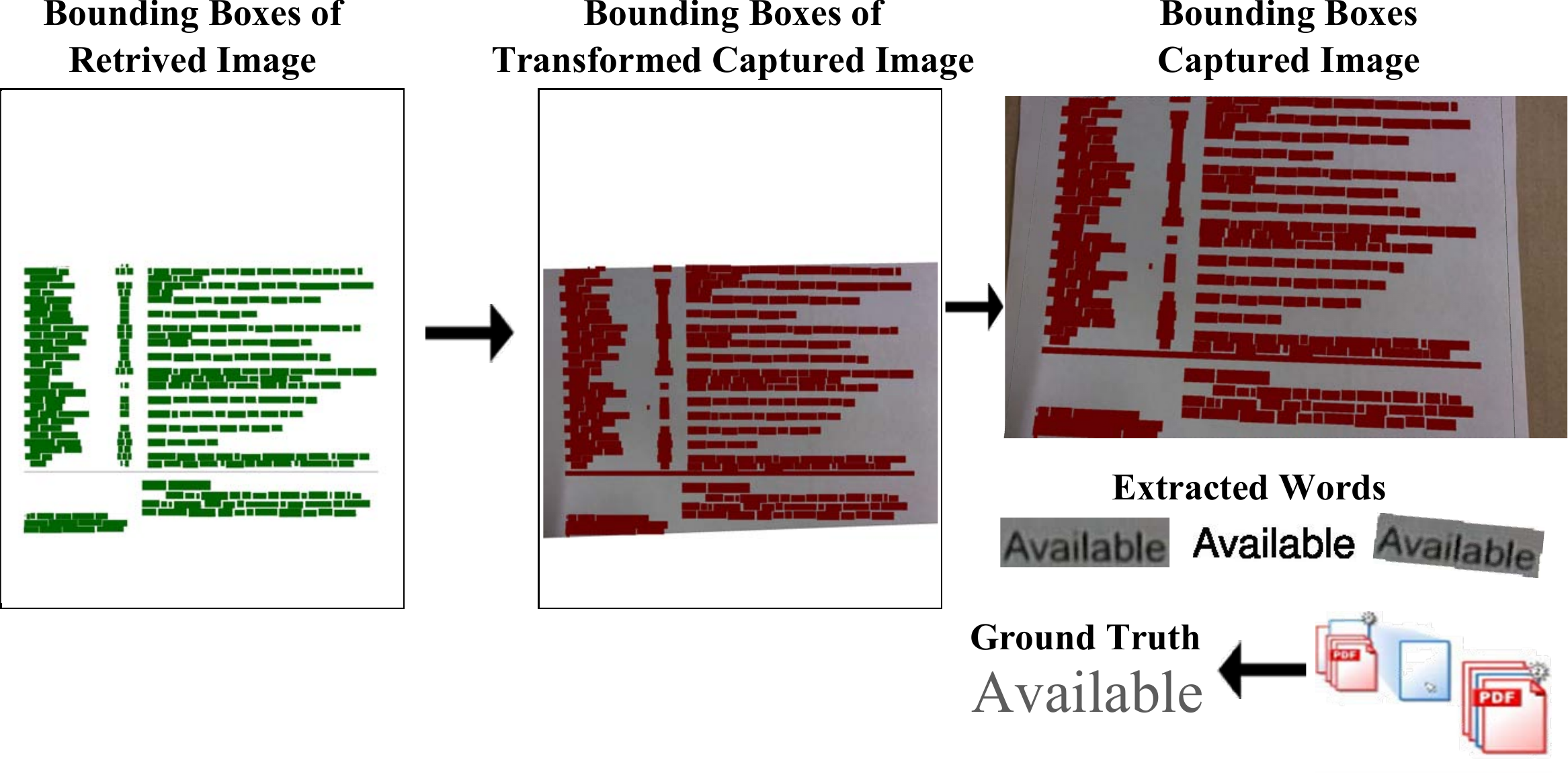}
	\caption{Words alignment and ground truth extraction}
 \label{fig:wlmatching}
\end{figure}

To find such word regions, the image is converted into word blocks by performing Gaussian smoothing on both the transformed captured image and the cropped electronic image. 
Bounding boxes are extracted from the smoothed images, where each box corresponds to a word in each image.
To find the corresponding words in both images, the distance between their centroids ($d_{\text{centroid}}$) and width ($d_{\text{width}}$) is computed.
The distance between centroids and width of bounding boxes is computed using the following equations.\\
\begin{equation} 
d_{\text{centroid}}=\sqrt{(\overline{x}_{\text{capt}}-\overline{x}_{\text{ret}})^2+(\overline{x}_{\text{capt}}-\overline{y}_{\text{ret}})^2} < \theta_{c}
\end{equation}

\begin{equation} 
d_{\text{width}}=\sqrt{(w_{\text{capt}}-w_{\text{ret}})^2} < \theta_{w}
\end{equation}

$(\overline{x}_{\text{capt}},\overline{y}_{\text{capt}}), w_{\text{capt}} $ and $(\overline{x}_{\text{ret}},\overline{y}_{\text{ret}}), w_{\text{ret}}$ refer to centroids and width of bounding boxes in the normalized/transformed captured and the cropped electronic image.
All of the boxes for which $d_{\text{centroid}}$ and $d_{\text{width}}$ are less than $\theta_{c}$ and $\theta_{w}$ respectively, are considered as boxes for the same word in both the images.
Here, $\theta_{c}$ and $\theta_{w}$ refer to the bounding box distance thresholds for centroid and width, respectively.\newline
We have used $\theta_{d}=5$ and $\theta_{w}=5$ pixels. This means if two boxes are almost at the same position in both images and their width is also almost the same, then they correspond to the same word in both images. 
All of the bounding boxes satisfying the criteria of Eqs. $(1)$ and $(2)$ are used to crop words from their respective images where no Gaussian smoothing is performed.
This results in two images for each word, i.e., the word image from the electronic document image (we call it ground truth image) and the word image from the transformed captured image. 

The word extracted from the transformed/normalized captured image is already normalized in terms of rotation, scale, and skew which were present in the originally captured image.
However, the original image with transformations and distortions is of main interest as it can be used for training of systems insensitive to different transformation.
To get the original image, inverse transformation is performed on the bounding boxes satisfying criteria set in Equations $(1)$ and $(2)$ in order to map them into the space of the original captured image containing different perspective distortions.
The boxes' dimensions after inverse transformation are then used to crop the corresponding words from original captured image. 
Finally, we have three different images for a word, i.e., from the electronic document image, from the transformed captured image, and the original captured image.
Note that the word images extracted from an electronic document are only an add-on, and have nothing to do with the camera-captured document.

Once these images are extracted, the next step is to associate them to their ground truth. 
To extract the text, we used the bounding box information of the word image from electronic/ground truth image (as this image was rendered from the PDF file) and extract text from the PDF for the bounding box. 
This extracted text is then saved as text file along with the word images.\\

To further extract characters from the word images, character bounding box information is used from PDF file of the retrieved document. 
In a PDF file, we have information about the bounding box of each character. Using this information, bounding boxes of characters in words satisfying the criteria of Eqs. $(1)$ and $(2)$ are extracted. 
These bounding boxes along with transformation parameters are then used for extracting character images from the original and the normalized/transformed captured images.
The text for each character is also saved along with each image.

Finally, we have characters extracted from the captured image and the normalized captured image.
Figure~\ref{fig:characters} and~\ref{fig:extractedwords} show the extracted characters and words images. 
\begin{figure}[t!]
  \centering  
	\includegraphics[width=\columnwidth]{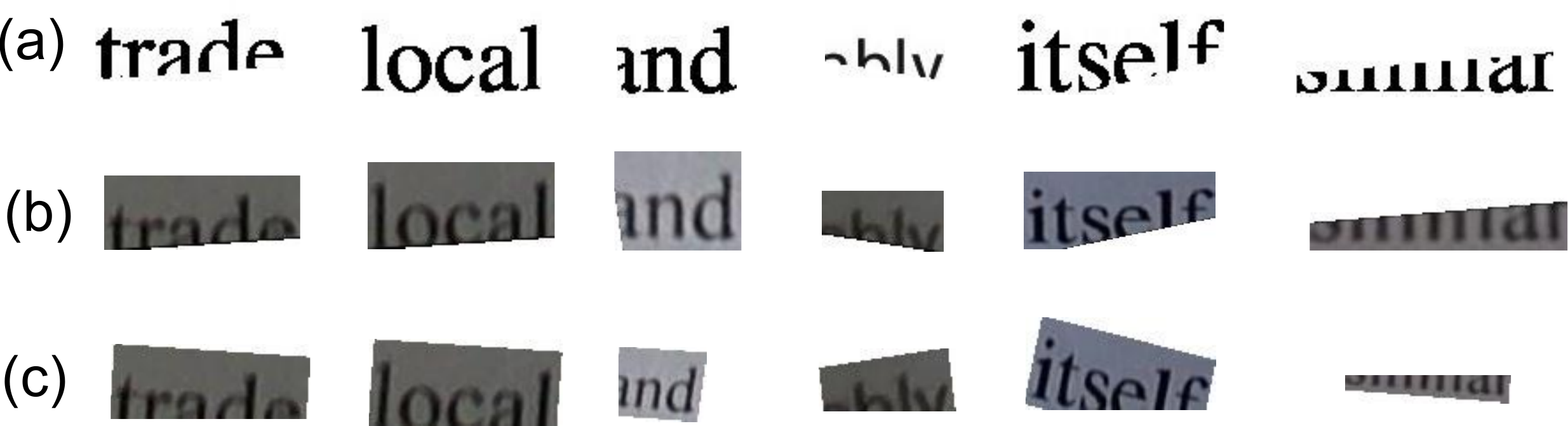}
	\caption{Words on border from (a) Retrieved image, (b) Normalized captured image, (c) Captured image}
 \label{fig:cornerwords}
\end{figure}
\subsection{Special cases}
As mentioned earlier, it is possible that a camera-captured image contain only a part of a document. 
Therefore, the region of interest could be any irregular polygon.
Figure~\ref{fig:plmatching} shows the estimated irregular polygon in green color.
Due to this, the characters and words that occur near or at the border of this region are partially missing. Figure~\ref{fig:cornerwords} shows some example words which occur at border of different camera-captured images.
These words, if included directly in the dataset, can cause problems during training, e.g., if a dot of an \textit{i} is missing then in some fonts it looks like 1 which can increase confusion between different characters.
To handle this problem, all the words and characters that occur near border are marked. 
This allows separating these words so that they can be handled separately if included in training.

\subsection{Cost analysis: Human vs. Automatic Method}\label{sec:costanalysis}
To get a quantitative measure and to find effectiveness and efficiency of the proposed method, cost analysis between human and the proposed ground truth generation method is performed. Ten documents, captured using camera, were given to a person to perform word and character level labeling.
The same documents were given as an input to the proposed system. 
The person performing labeling task took $670$ minutes to label these documents. 
To crop words from the document it took additional $940$ minutes. 
In total the person took $1610$ minutes to extract words and label them. 
On the other hand, for the same documents, our system was able to extract all words and character images with their ground truth, and normalized images (where they are corrected for different perspective distortion) in less than $2$ minutes. 
This means that the presented automatic method is almost 800 times faster than human. It also confirms the claim that it is not possible to build very large-scale datasets by manual labeling due to extensive cost and time. 
With the presented approach, it is possible to build large-scale datasets in very short time. The only thing, which needs to do, is document capturing. 
The rest is managed by the method itself.

\begin{figure}[t!]
  \centering  
	\includegraphics[width=\columnwidth]{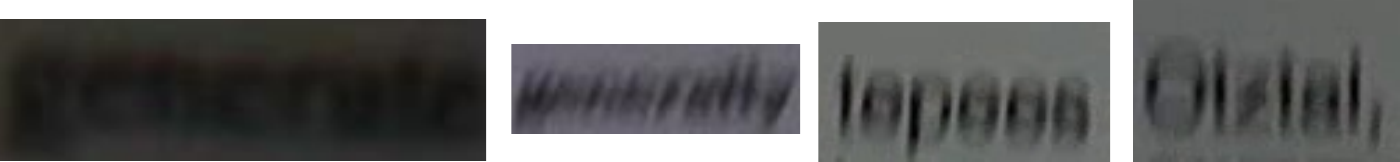}
	\caption{Words where human faced difficulty in labeling}
 \label{fig:exampledifficultlabel}
\end{figure}

Another important benefit of the proposed method over human is that the presented method is able to assign ground truth to even severely distorted images where even humans were unable to understand the content.
Figure~\ref{fig:exampledifficultlabel} shows example words where the human had difficulty in labeling but were successfully labeled by the proposed method.

\subsection{Evaluation of Automatic Ground Truth Generation Method}\label{sec:evaluation}
To evaluate the precision and prove that the proposed method is generic, two datasets are generated: one in English and other in Russian. 
The dataset in English consist of one million word and ten million character images. The dataset in Russian contains approximately $100,000$ word and $500,000$ character images.
Documents used for generation of these datasets are diverse and include books, magazine, articles, etc. 
These documents are captured using different cameras ranging from high-end cameras to normal web-cams. 

Manual evaluation is performed to check correctness and quality of the generated ground truth.
Out of the generated dataset, $50,000$ samples were randomly selected for evaluation.
One person has manually inspected all of these samples to find out errors.
This manual check shows that more than $99.98$\% of the extracted samples are correct in term of ground truth as well as the extracted image.
A word or character is referred to as correct if and only if the content in cropped word from electronic image, the transformed captured image, the original captured image, and the ground truth text corresponding to these images are the same.
While evaluating, it is also taken into account that each image should exactly contain the same information.
The 0.02\% error is due to the problem faced by the ground-truth method in labeling very small font size (for instance 6 size) words having punctuations at the end.
In addition to camera-captured images, the proposed method is also tested on scanned images, where it has also achieved an accuracy of more than $99.99$\%.
This means that almost all of the images are correctly labeled.

\section{Camera-Captured Characters and Word images (C$^{3}$W\textit{i}) Dataset}\label{dataset}
A novel dataset of camera-captured character and word images is also introduced in this paper. 
This dataset is generated using the method proposed in Section~\ref{sec:Methodology}. 
It\footnote{If the paper is accepted, the dataset will be publicly available} contains one million words and ten million character images extracted from different text documents.
These characters and words are extracted from diverse collection of documents including conference proceedings, books, magazines, articles, and newspapers. 
The documents are first captured using three different cameras ranging from normal web cams to high-end cameras, having resolution from two megapixels to eight megapixels. In addition, documents are captured under varying lighting conditions, with different focus, orientation, perspective distortions, and out of focus settings.
Figure~\ref{fig:samples} shows sample documents captured using different cameras and in different settings.
Captured documents are then passed to the automatic ground truth generation method, which extracts word and character images from the camera-captured documents and attach ground truth information from PDF file.

Each word in the dataset has the following three images:
\begin{figure}[t!]
  \centering  
	\includegraphics[width=\columnwidth]{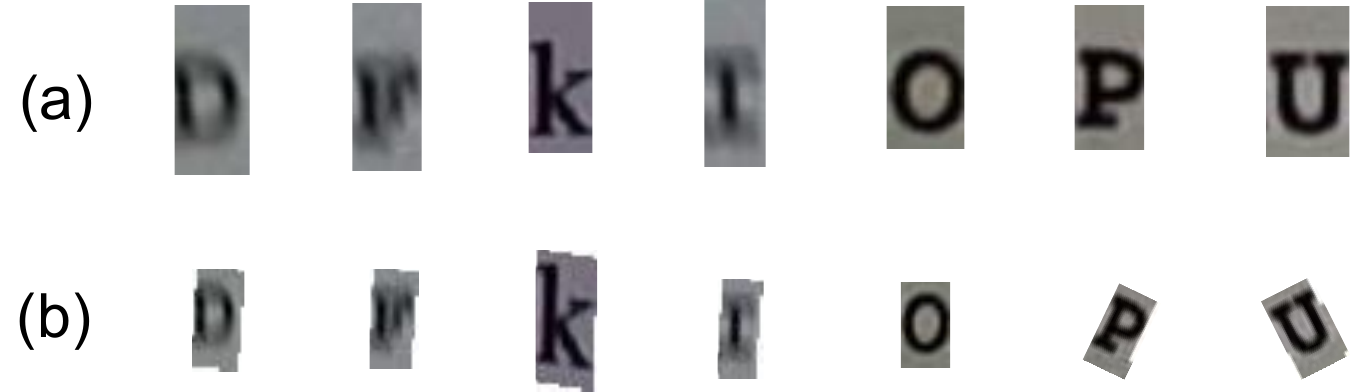}
	\caption{Extracted characters from (a) Normalized captured image, (b) Captured image}
 \label{fig:characters}
\end{figure}

\begin{figure*}[t!]
  \centering  
	\includegraphics[width=\textwidth]{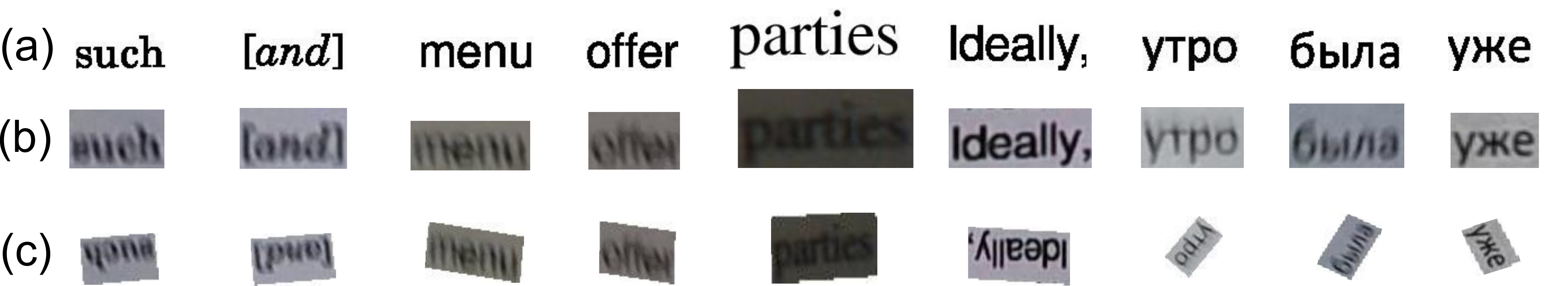}
	\caption{Word sample from an automatically generated dataset. (a) Ground truth image, (b) Normalized camera-captured image (c) camera-captured image with distortions}
 \label{fig:extractedwords}
\end{figure*}
\begin{itemize}
	\item Ground truth word image: This is a clean word image extracted from the electronic version (ground truth) of the camera-captured document. Figure~\ref{fig:extractedwords} (a) shows example ground truth word images extracted by the ground truth generation method.
	\item Normalized word image: This image is extracted from normalized camera-captured document. This means that it is corrected in terms of perspective distortion, but still contains qualitative distortions like blur, varying lighting condition, etc. Figure~\ref{fig:extractedwords} (b) shows example normalized word images extracted by the ground truth generation method.
	\item Original camera-captured word image: This image is extracted from the original camera-captured document. It contains various distortions specific to camera-captured images, e.g., perspective distortion, blur, and varying lighting condition. Figure~\ref{fig:extractedwords} (c) shows example camera-captured word images extracted by the proposed ground truth generation method.

\end{itemize}

In addition to these images, a text ground truth is also attached with a word, which contains actual text present in the camera-captured image.

Similarly, each character in the dataset has two images:
\begin{itemize}
	\item Normalized character image: This image is extracted from normalized camera-captured document. This means that it is corrected in terms of perspective distortion, but still contains qualitative distortions like blur, varying lighting condition, etc. Figure~\ref{fig:characters} (a) shows the example normalized character images extracted by the ground truth generation method.
	\item Original camera-captured character image: This image is extracted from the original camera-captured document. It contains various distortions specific to camera-captured images, e.g., perspective distortion, blur, and varying lighting condition. Figure~\ref{fig:characters} (b) shows the example camera-captured character images extracted by the ground truth generation method.

\end{itemize}

For each character image, a ground truth file (containing text) is also associated, which contains characters present in an image.
 
In total, the dataset contains three million word images along with one million word ground truth text files and twenty million character images with ten million ground truth files.

The Dataset is divided into training, validation, and test set. 
Training set includes $600,000$ words and six million characters. This means that $60\%$ of the dataset is available for training. 
The validation set includes $100,000$ words (one million characters).
The test set includes the remaining $300,000$ words and three million character images. 

\section{Neural Network Recognizer:\\ The Proposed Character Recognition System}~\label{sec:proposedmethod:nn}

In addition to automatic ground truth generation method and C$^{3}$W\textit{i} datatset, this paper also presents a character recognition system for camera-captured document images. The proposed recognition system is based on Long Short Term Memory (LSTM), which is a modified form of Recurrent Neural Network (RNN). 

Although RNN performs very well in the sequence classification tasks, it suffers from the vanishing gradient problem. 
The problem arises when the error signal flowing backwards for the weight correction vanish and thus are unable to model long-term dependencies/contextual information.
In LSTM, the vanishing gradient problem does not exist and, therefore, LSTM can model contextual information very well.

Another reason for proposing an LSTM based recognizer is that they are able to learn from large unsegmented data and incorporate contextual information. This contextual information is very important in recognition. This means that while recognizing a character it incorporates the information available before the character.

\begin{figure}[t!]
  \centering  
	\includegraphics[width=0.8\columnwidth]{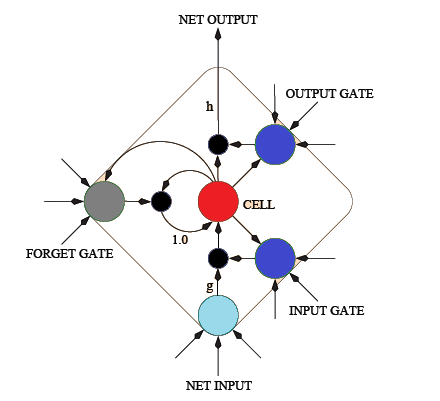}
	\caption{LSTM memory block~\cite{marcus-tpami08}}
 \label{fig:lstm-cell}
\end{figure}

The structure of the LSTM cell can be visualized as in Figure~\ref{fig:lstm-cell} and simplified version is mathematically expressed in Eqs.~(\ref{eq:lstm1}), ~(\ref{eq:lstm2}) and ~(\ref{eq:lstm3}). Here, the traditional RNN unit is modified and multiplicative gates, namely input ($I$), output ($O$), and forget gates ($F$), are added.
The state of the LSTM cell is preserved internally. The reset operation of the internal memory state is protected with forget gate which determines the reset of memory based on the contextual information.
\begin{equation} \label{eq:lstm1}
\centering
\begin{pmatrix}
I\\ F\\ C\\ O
\end{pmatrix} = f(W.X^t+W.H_{d}^{t-1}+W.S_{c,d}^{t-1})
\end{equation}

\begin{equation}\label{eq:lstm2}
S_{c}^{t}=I.C+S_{c,d}^{t-1}.F_d
\end{equation}
\begin{equation}\label{eq:lstm3}
H^t = O.f(S_{c}^{t})
\end{equation}
The input, forget, and output gates are denoted by $I$, $F$, and $O$ respectively. The $t$ denotes the time-step and in our case, a pixel or a block. The number of recurrent connections are
equivalent to the dimensions which are represented by $d$. It is to be noted that for exploiting the temporal cues for recognition, the word images are scanned in 1D. So, for the equations mentioned above, the value of $d$ is $1$.

In offline data it is possible to use both the past and the future contextual information by scanning them from both direction, i.e., left-to-right and right-to-left. 
An augmentation of the one directional LSTM is the bidirectional long short term memory (BLSTM)~\cite{Graves-2008-thesis,marcus-tpami08}. 
In the proposed method, we used BLSTM where each word image is scanned from left to right and from right to left. This is accomplished by having two one directional LSTM but the scanning is done in different directions. Both of these hidden layers are connected to output layer for providing the context information from both the past and the future.
In this way at a current time step, while predicting a label, we would be able to have the context both from the past and from the future. 

Some earlier researchers, like Bissacco et al.~\cite{PhotoOCR} used fully connected neural networks. However, segmented characters are required to train their system. Furthermore, to incorporate contextual information they used language modeling.
Although in the proposed dataset, we have provided character data as well but we are still using unsegmented data. This is because, with unsegmented data, LSTM is able to automatically learn the context.  
Furthermore, segmentation of data itself can lead to under and/or over segmentation, which can lead to problems during training, whereas in unsegmented data this problem simply does not exist. 
RNNs also require pre-segmented data where target has to be specified at each time step for the prediction purpose. This is generally not possible in unsegmented sequential data where the output labels are not aligned with the input sequence. To overcome this difficulty and to process the unsegmented data Connectionist Temporal Classification (CTC) has been used as an output layer of LSTM~\cite{Graves06connectionisttemporal}. The algorithm used in CTC is forward backward algorithm, which requires the labels to be presented in the same order as they appear in the unsegmented input sequence. The combination of LSTM and CTC yielded the state-of-the-art results in handwriting analysis~\cite{marcus-tpami08}, printed character recognition~\cite{breuel-2013}, and speech recognition~\cite{Graves-speech13,Graves-speech13a}.
\begin{figure}[t!]
  \centering  
	\includegraphics[width=\columnwidth]{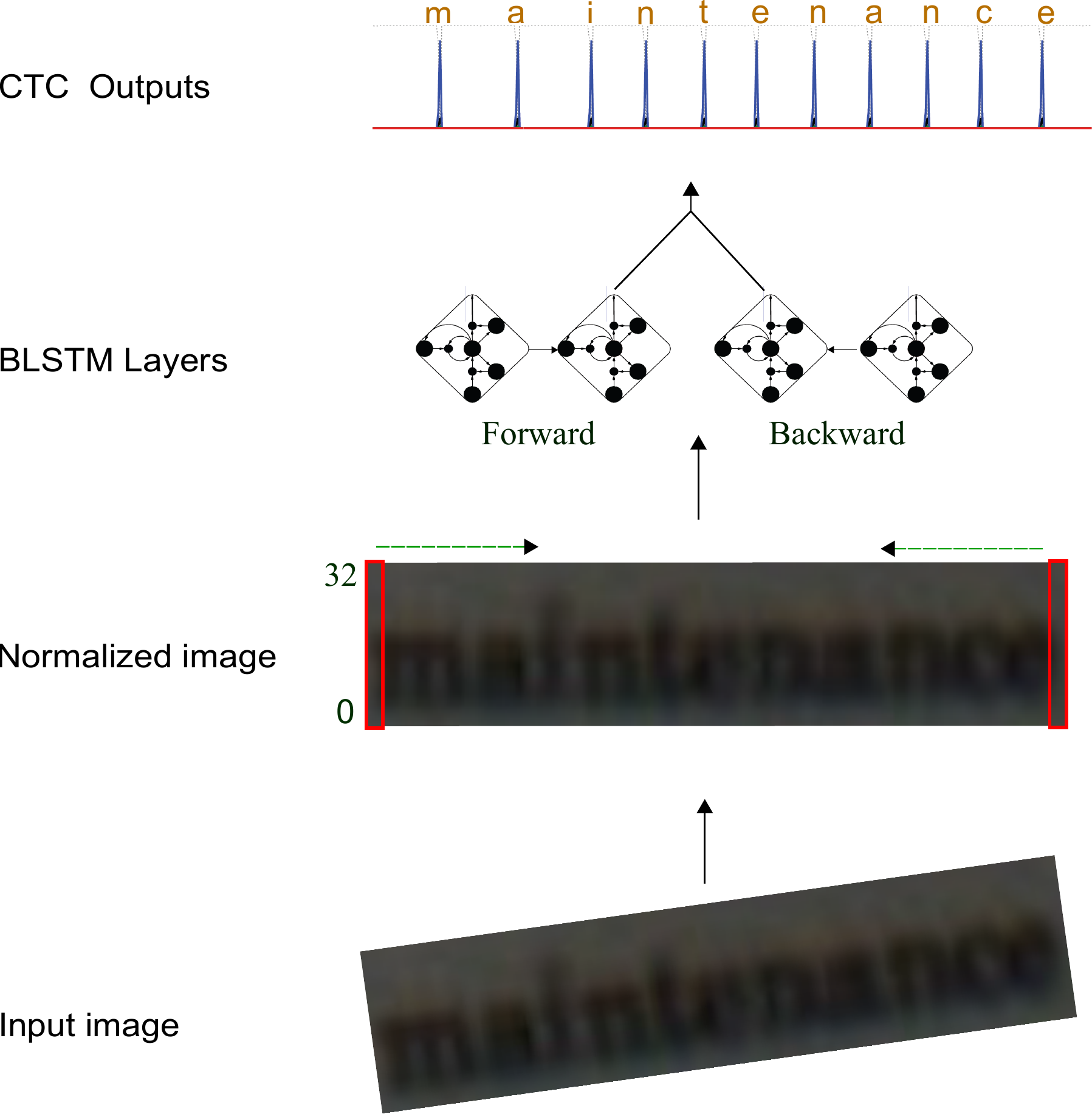}
	\caption{Architecture of LSTM based recognizer}
 \label{fig:lstm-flow}
\end{figure}

In the proposed system, we used BLSTM architecture with CTC to design the system for recognition of camera-captured document images. 
BLSTM scans input from both directions and learn by incorporating context into account.
Unsegmented word images are given as an input to BLSTM. 
Contrary to Bissacco et al.~\cite{PhotoOCR}, where the histogram of oriented gradients (HOG) features are used, the proposed method takes raw pixel values as input for LSTM and no sophisticated features extraction is performed.
The motivation behind raw pixels is to avoid handcrafted features and to present LSTM with the complete information so that it can detect and learn relevant features/information automatically. 
Geometric corrections, e.g., rotation, skew, and slant correction is performed on input images. 
Furthermore, height normalization is performed on word images that are already corrected in terms of geometric distortions. 
Each word image is rescaled to the fixed height of $32$ pixels. 
The normalized image is scanned from right to left with a window of size $32 X 1$ pixels.
This scanning results into a sequence that is fed into BLSTM. 
The complete LSTM based recognition system is shown is Figure~\ref{fig:lstm-flow}. 
The output of BLSTM hidden layers is fed to the CTC output layer, which produces a probability distribution over character transcriptions.

\begin{figure}[t!]
  \centering  
	\includegraphics[width=\columnwidth]{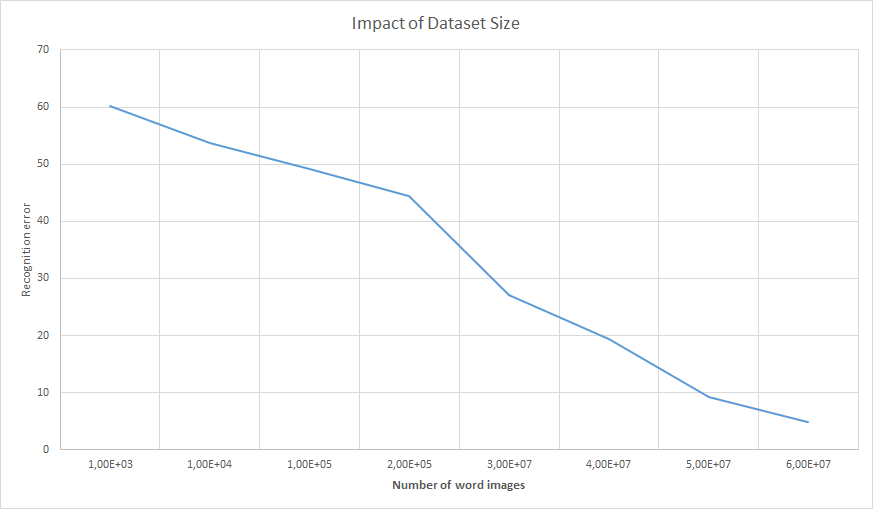}
	\caption{Impact of dataset size on recognition error}
 \label{fig:influence}
\end{figure}

Note that various sophisticated classifiers, like SVM, cannot be used with large datasets as they can be expensive in time and space. However, LSTM is able to handle and learn from large datasets. 
Figure~\ref{fig:influence} shows the impact of increase in dataset size on the overall recognition error in the presented system. It can be seen that with the increase in dataset size, overall recognition error drops.
The trend in Fig.~\ref{fig:influence} also shows the importance of having large datasets which can be generated using the automatic ground truth generation method presented in this paper. 
As this method (explianed in Section~\ref{sec:Methodology}) is language independent, we can build very large datasets for different languages, which in turn will result into accurate OCRs for different languages.

\subsection{Parameter Selection}
In LSTM, the hidden layer size is an important parameter. The size of hidden layer is directly proportional to training time.
This means that increasing the number of hidden units increases the time required to train the network.
In addition to time, hidden layer size also affects the learning of network. 
A network with few numbers of hidden units results in high recognition error. Whereas, a network with large number of hidden units converges to an over-fitted network.
To select an appropriate number of hidden layers, we trained multiple networks with different hidden units configuration including $40, 60, 80, 100, 120, 140$. 
We selected network with $100$ layers because after $100$ error rate on the validation set started increasing. 

The training and validation set of C$^3$W\textit{i} consisting of $600,000$ and $100,000$  images, respectively, are used to train the network with hidden size of $100$, momentum of $0.9$, and learning rate of $0.0001$.

\section{Performance Evaluation of the Proposed and Existing OCRs}\label{expeval}

The aim of this evaluation is to gauge the performance and behavior of existing and proposed character recognition systems on camera-captured document images.
To do so, we used the C$^3$W\textit{i} dataset, which is generated using the method proposed in Section~\ref{sec:Methodology}.
We trained our method on the training set of C$^3$W\textit{i} dataset.
ABBYY and Tesseract already claim to support camera based OCRs~\cite{ABBYY-online,PhotoOCR}.  
As mentioned in Section~\ref{dataset}, each word in the dataset has three different images and a text ground truth file, i.e., original camera-captured word image, normalized camera-captured word image, and ground truth image. 
To have a thorough and in-depth evaluation of OCRs, two different experiments are performed.

\begin{itemize}	
	\item \textbf{Experiment 1:} Normalized version of camera-captured word images where original camera-captured images are normalized in terms of perspective distortions (Figure~\ref{fig:extractedwords}(b)), are passed to ABBYY, Tesseract, and the proposed LSTM based character recognition system.
	Note that these images still contain qualitative distortions e.g., blur, and varying lighting.
\item\textbf{Experiment 2:} Ground truth word image extracted from the electronic version of captured document (Figure~\ref{fig:extractedwords}(a)), are passed to  ABBYY, Tesseract, and the proposed LSTM based character recognition system.
\end{itemize}

To find out the accuracy, a Levenshtein distance~\cite{Levenshtein} based accuracy measure is used.
This measure includes the number of insertions, deletion, and substitutions, which are necessary for converting a given string into another. 
Equation~\ref{acc} is used for measuring the accuracy.

	\begin{equation}
\footnotesize
	\label{acc}
Accuracy = 1-\frac{(\text{insertions} + \text{substitutions} +\text{deletions})}{\text{len(ground truth transcription)}} * 100
	\end{equation}

	\begin{table}[!ht]
\centering
\caption{Recognition accuracy of OCRs for different experiments.
}\label{tab:ocrs}
\begin{tabular}{|c|c|c|}
\hline
\textbf{OCR Name}&Experiment 1 &Experiment 2\\
\hline
\emph{Tesseract}&$50.44\%$&$94.77\%$\\
\emph{ABBYY FineReader}&$75.02\%$&$99.41\%$\\
\emph{Neural Network Based Recognizer}&$95.10\%$&$97.25\%$\\
\hline
\end{tabular}  
\end{table}

\begin{table*}[!ht]
\centering
\caption{Sample results for camera-captured words with distortions}\label{tab:samples}
\begin{tabular}{|c|c|c|c|c|c|}
\hline
 \textbf{Index} &\textbf{Sample Image} &\textbf{Ground Truth }& \textbf{Tesseract} &\textbf{ABBYY FineReader }&\textbf{Proposed Recognizer}\\
\hline
1&\emph{\includegraphics[width=0.18\textwidth,height=1cm,keepaspectratio]{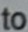}}&to&IO&to&to\\
\hline
2&\emph{\includegraphics[width=0.18\textwidth,height=1cm,keepaspectratio]{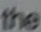}}&the& &the&thn\\
\hline
3&\emph{\includegraphics[width=0.18\textwidth,height=1cm,keepaspectratio]{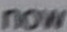}}&now&I&no &now\\
\hline
4&\emph{\includegraphics[width=0.18\textwidth,height=1cm,keepaspectratio]{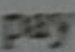}}&pay&'5& &py\\
\hline
5&\emph{\includegraphics[width=0.18\textwidth,height=1cm,keepaspectratio]{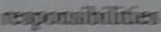}}&responsibilities&j& &responsibiites\\
\hline
6&\emph{\includegraphics[width=0.18\textwidth,height=1cm,keepaspectratio]{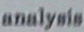}}&analysis&umnlym& HNMlyill«&annlysis\\
\hline
7&\emph{\includegraphics[width=0.18\textwidth,height=1cm,keepaspectratio]{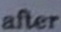}}&after&.& -îfU-r&after\\
\hline
8&\emph{\includegraphics[width=0.18\textwidth,height=1cm,keepaspectratio]{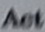}}&act& &Ai t&act\\
\hline
9&\emph{\includegraphics[width=0.18\textwidth,height=1cm,keepaspectratio]{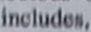}}&includes,&Ingludul&Include*,&includes,\\
\hline
10&\emph{\includegraphics[width=0.18\textwidth,height=1cm,keepaspectratio]{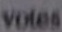}}&votes&Will&Virilit&voes\\
\hline
11&\emph{\includegraphics[width=0.18\textwidth,height=1cm,keepaspectratio]{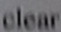}}&clear&vlvur&t IHM&clear\\
\hline
12&\emph{\includegraphics[width=0.18\textwidth,height=1cm,keepaspectratio]{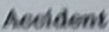}}&Accident&Maiden!&&Aceident\\
\hline
13&\emph{\includegraphics[width=0.18\textwidth,height=1cm,keepaspectratio]{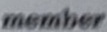}}&member&.&&meember\\
\hline
14&\emph{\includegraphics[width=0.18\textwidth,height=1cm,keepaspectratio]{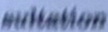}}&situation&mum&&sltstion\\
\hline
15&\emph{\includegraphics[width=0.18\textwidth,height=1cm,keepaspectratio]{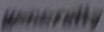}}&generally& &&genray\\
\hline
16&\emph{\includegraphics[width=0.18\textwidth,height=1cm,keepaspectratio]{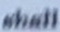}}&shall&WNIW&•».»II&adad\\
\hline
17&\emph{\includegraphics[width=0.18\textwidth,height=1cm,keepaspectratio]{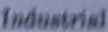}}&Industrial&[Mandi&&Industril\\

\hline
\end{tabular}  
\end{table*}

Table~\ref{tab:ocrs} shows the accuracy for all the experiments. 
The results of Experiment 1 shows that both Commercial (ABBYY) as well as open source (Tesseract) OCRs fail severely when applied on camera-captured word images. 
The main reason for this failure is the presence of distortions specific to camera-captured images, e.g., blur, and varying lighting conditions. It is to be noted that images used in this experiment are normalized for geometric distortions like, rotation, skew, slant, and perspective distortion.
Even in the absence of perspective distortions, existing OCRs fail. This shows that quality related distortions, e.g., blur and varying lighting have strong impact on recognition in existing OCRs.
Table~\ref{tab:samples} shows some sample images along with OCR results of different systems.

To show that OCRs are really working on word images, Experiment $2$ was performed. In this experiment, clean word images extracted from electronic versions of camera-captured documents are used. These documents do not contain any geometric or qualitative distortion.
These clean ground truth word images are passed to the OCRs.
All of the OCRs performed well and achieved very high accuracies.
In addition, our proposed system performed even better than the Tesseract and achieved a performance close to ABBYY. 
It is to be noted that our system was only trained on camera-captured images, and no clean image was used for training. 
The result of this experiment shows that if trained on degraded images, the proposed system can recognize both degraded as well as clean images. 
However, the other way around is not true, as existing OCRs fail on camera-captured images but perform well on clean word images.

The analysis of the experiments shows that the existing OCRs, which already get very high accuracies on scanned documents, i.e., 99.94\%, have a limited performance on camera-captured document images with the best recognition accuracy of $75.02$\% in case of commercial OCR (ABBYY) and $50.44$\% in case of open source OCR (Tesseract). On deeper examination, it is further observed that main reason for failure of existing OCRs is not the perspective distortion, but the qualitative distortions. 

To confirm our findings, we performed another experiment where images with blur and bad lighting were presented to all recognizers. 
These results are summarized in Table~\ref{tab:ocrs-blur}.
Analysis of results in Table~\ref{tab:ocrs-blur} confirms that both commercial (ABBY FineReader) as well as open source (Tesseract) OCRs fail severely on images with blur and bad lighting conditions. 
On the other hand, they are performing well on clean images, regardless of camera-captured or scanned images.
The main reason of this failure is qualitative distortions (i.e., blurring and varying lighting conditions), especially, if images are of low contrast, almost all existing OCRs fail to recognize text.
While the proposed LSTM based recognizer is able to recognize them with an accuracy of 86.8\%.

This effect could be seen in Table~\ref{tab:samples}, where the outputs of existing OCRs are not even close to the ground truth. This is because most of these systems are using binarization before recognition. If low contrast images are not binarized properly, there will be too much noise and loss of information, which would result in miss-classification. While the proposed LSTM based recognizer performs reasonably well. It generates outputs close to the ground truth, even for those cases where it is difficult for humans to understand the content, e.g., row $14$ and $15$ in Table~\ref{tab:samples}.

\begin{table}[!ht]
\centering
\caption{Recognition accuracy of OCRs on only blur and varying lighting images.
}\label{tab:ocrs-blur}
\begin{tabular}{|c|c|}
\hline
OCR&Accuracy\\
\hline
Tesseract & 18.1\\
\hline
ABBYY FineReader& 19.57\\
\hline
Proposed System& 86.8\\
\hline
\hline
\end{tabular}  
\end{table}

Furthermore, note that all the results of the proposed character recognition system are achieved without any language modeling. 
Analysis of results reveals that there are few mistakes, which can be easily avoided by incorporating language modeling. 
For example in Table~\ref{tab:samples} the word ``voes'' can be easily corrected to ``votes''.

\section{Conclusion}\label{sec:conclusion}
In this paper, we proposed a novel and generic method for automatic ground truth generation of camera-captured/scanned document images.
The proposed method is capable of labeling and generating large-scale datasets very quickly.
It is fully automatic and does not require any human intervention for labeling. 
Evaluation of the sample from generated datasets shows that our system can be successfully applied to generate very large-scale datasets automatically, which is not possible via manual labeling.
While comparing the proposed ground-truth generation method with humans, it was revealed that the proposed method is able to label even those words where humans face difficulty even in reading, due to bad lighting condition and/or blur in the image.
The proposed method is generic as it can be used for generation of dataset in different languages (English, Russian, etc.).
Furthermore, it is not limited to camera-captured documents and can be applied to scanned images.

In addition to a novel automatic ground truth generation method, a novel dataset of camera-captured documents consisting of one million words and ten million labeled character images is also proposed.
The proposed dataset can be used for training and testing of OCRs for camera-captured documents.
Furthermore, along with the dataset, we also proposed an LSTM based character recognition system for camera-captured document images. 
The proposed character recognition system is able to learn from large datasets and therefore trained on C$^3W$\textit{i} dataset.
Various benchmark tests were performed using the proposed C$^3W$\textit{i} dataset to evaluate the performance of different open source (Tesseract~\cite{Tesseract1}), commercial (ABBYY~\cite{ABBYY,Tesseract1}), as well as proposed LSTM based character recognition system.
Evaluation results show that both commercial (ABBYY with an accuracy of 75.02\%) and open source (Tesseract with an accuracy of 75.02\%) OCRs fail on camera-captured documents, especially due to qualitative distortions which are quite common in camera-captured documents. Whereas, the proposed character recognition system is able to deal with severly blurred and bad lighting images with an overall accuracy of 95.10\%.

In the future, we  plan to build dataset for different languages, including Japanese, Arabic, Urdu, and other Indic scripts, as there is already a strong demand for OCR of different languages e.g., Japanese~\cite{JapaneseOCR}, Arabic~\cite{ArabicOCR}, Indic scipts~\cite{IndicOCR}, Urdu~\cite{UrduOCR},  etc., and each one needs a different dataset specifically built for that language.
Furthermore, we are also planning to use the proposed dataset for domain adaptation. This means that training a model on C$^3W\textit{i}$ dataset with the aim to make it working on natural scene images. 

\ifCLASSOPTIONcompsoc
  \section*{Acknowledgments}
\else

  \section*{Acknowledgment}

\fi
This work is supported in part by CREST and JSPS Grant-in-Aid for Scientific Research (A)(25240028).

\ifCLASSOPTIONcaptionsoff
  \newpage
\fi

\bibliographystyle{IEEEtran}
\bibliography{Bibliography}
%
\begin{IEEEbiography}[{\includegraphics[width=1.1in,height=1.25in,clip]{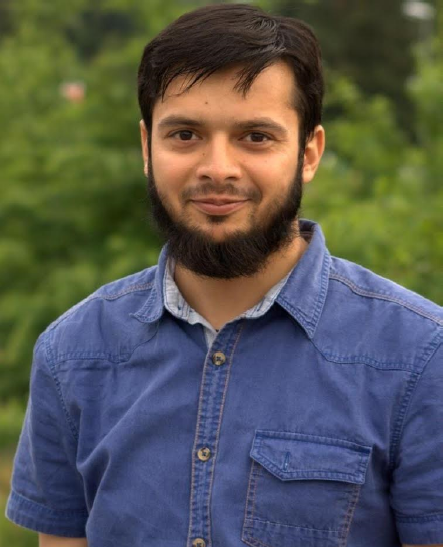}}]{Sheraz Ahmed} received   his  Masters degree  (from  the  Technische  Universitaet Kaiserslautern, Germany) in Computer Science. Over the last few years, he has primarily worked for development of various systems for  information  segmentation  in  document images. Recently he completed his PhD
in the German Research Center for Artificial Intelligence, Germany, under the supervision
of Prof. Dr. Prof. h.c. Andreas Dengel and Prof. Dr. habil. Marcus Liwicki. His PhD topic is
Generic Methods for Information Segmentation in Document Images.
His research interest includes document understanding, generic
segmentation framework for documents, gesture recognition, pattern recognition, data mining, anomaly detection, and natural language processing. He has more than $18$
publications on the said and related topics including three journal papers and two book chapters. He is a frequent reviewer of various journals and conferences including
Patter Recognition Letters, Neural Computing and Applications, IJDAR,
ICDAR, ICFHR, DAS, and so on. From October $2012$ to April $2013$ he visited Osaka Prefecture University (Osaka, Japan) as a research fellow, supported by the Japanese Society for the Promotion of Science and from September $2014$ to November $2014$ he visited University of
Western Australia (Perth, Australia) as a research fellow, supported by
the DAAD, Germany and $Go-8$, Australia
\end{IEEEbiography}

\begin{IEEEbiography}[{\includegraphics[width=1.1in,height=1.25in,clip]{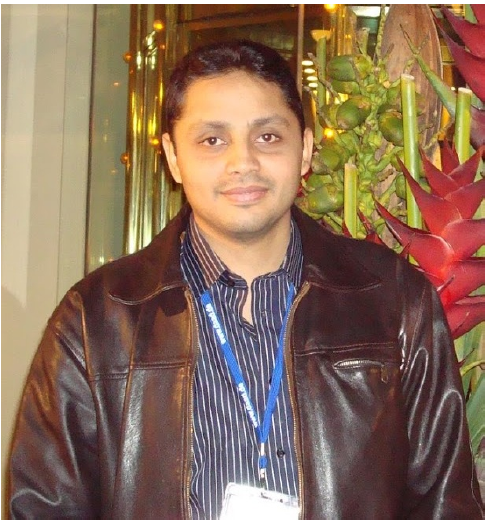}}]{Muhammad Imran Malik} received both his Bachelors (from Pakistan) and Masters (from the Technische Universitaet Kaiserslautern, Germany) degrees in Computer Science. In Bachelors thesis, he
worked in the domains of real time object detection and image enhancement. In Masters thesis,
he primarily worked for development of various systems for signature identification and verification. Recently, he completed his PhD in the German Research Center for Artificial Intelligence, Germany, under the supervision of Prof. Dr. Prof. h.c. Andreas Dengel
and PD Dr. habil. Marcus Liwicki. His PhD topic is Automated Forensic Handwriting Analysis on which he has been focusing from both the perspectives of Forensic Handwriting Examiners (FHEs) and Pattern Recognition (PR) researchers. He has more than 25 publications on the
said and related topics including two journal papers.\end{IEEEbiography}
\begin{IEEEbiography}[{\includegraphics[width=1.1in,height=1.25in,clip]{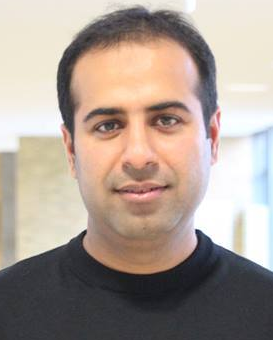}}]{Muhammad Zeshan Afzal} received his Masters degree (University of Saarland, Germany) in Visual Computing in 2010. Currently he is a PhD candidate in University of Technology, Kaiserslautern, Germany. His research interests include generic segmentation framework for natural, document and, medical images, scene text detection and recognition, on-line and off-line gesture recognition, numerics for tensor valued images, pattern recognition with special interest in recurrent neural network for sequence processing applied to images and videos. 
He received the gold medal for the best graduating student in Computer Science from IUB Pakistan in 2002 and secured a DAAD(Germany) fellowship in 2007. He is a member of IAPR.
\end{IEEEbiography}

\begin{IEEEbiography}[{\includegraphics[width=1.1in,height=1.25in,clip]{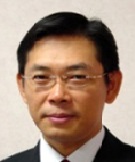}}]{Koichi Kise}
received B.E., M.E. and Ph.D. degrees in communication engineering from Osaka University, Osaka, Japan in 1986, 1988 and 1991, respectively. From 2000 to 2001, he was a visiting professor at German Research Center for Artificial Intelligence (DFKI), Germany.
He is now a Professor of the Department of Computer Science and Intelligent Systems, and the director of the Institute of Document Analysis and Knowledge Science (IDAKS), Osaka Prefecture University, Japan. He received awards including the best paper award of IEICE in 2008, the IAPR/ICDAR best paper awards in 2007 and 2013, the IAPR Nakano award in 2010, the ICFHR best paper award in 2010 and the ACPR best paper award in 2011. He works as the chair of the IAPR technical committee 11 (reading systems), a member of the IAPR conferences and meetings committee, and an editor-in-chief of the international journal of document analysis and recognition.
His major research activities are in analysis, recognition and retrieval of documents, images and activities.
He is a member of IEEE, ACM, IPSJ, IEEJ, ANLP and HIS.
\end{IEEEbiography}
\begin{IEEEbiography}[{\includegraphics[width=1.1in,height=1.25in,clip]{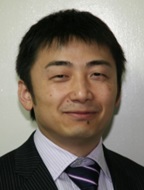}}]{Masakazu Iwamura}
received the B.E., M.E., and Ph.D degrees in communication engineering from Tohoku University, Japan, in 1998, 2000 and 2003, respectively. He is an associate professor of the Department of Computer Science and Intelligent Systems, Osaka Prefecture University. He received awards including the IAPR/ICDAR Young Investigator Award in 2011, the best paper award of IEICE in 2008, the IAPR/ICDAR best paper awards in 2007, the IAPR Nakano award in 2010, and the ICFHR best paper award in 2010. He works as the webserver of the IAPR technical committee 11 (Reading Systems). His research interests include statistical pattern recognition, character recognition, object recognition, document image retrieval and approximate nearest neighbor search.
\end{IEEEbiography}
\begin{IEEEbiography}[{\includegraphics[width=1.1in,height=1.25in,clip]{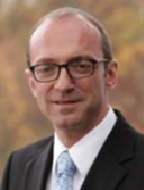}}]{Andreas Dengel}
is a Scientific Director at the German Research Center for Artificial Intelligence (DFKI GmbH) in Kaiserslautern. In 1993, he became a Professor at the Computer Science Department of the University of Kaiserslautern where he holds the chair “Knowledge-Based Systems” and since 2009 he is appointed Professor (Kyakuin) at the Department of Computer Science and Information Systems at the Osaka Prefecture University. He received his Diploma in CS from the University of Kaiserslautern and his PhD from the University of Stuttgart. He also worked at IBM, Siemens, and Xerox Parc. Andreas is member of several international advisory boards, chaired major international conferences, and founded several successful start-up companies. Moreover, he is co-editor of international computer science journals and has written or edited 12 books. He is author of more than 300 peer-reviewed scientific publications and supervised more than 170 PhD and master theses. Andreas is a IAPR Fellow and received prominent international awards. His main scientific emphasis is in the areas of Pattern Recognition, Document Understanding, Information Retrieval, Multimedia Mining, Semantic Technologies, and Social Media.
\end{IEEEbiography}
\begin{IEEEbiography}[{\includegraphics[width=1.1in,height=1.25in,clip]{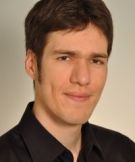}}]{Marcus Liwicki}
Marcus Liwicki received his PhD degree from the University of Bern, Switzerland, in
2007. Currently he is a senior researcher 
at the German Research Center for
Artificial Intelligence (DFKI) and Professor at Technische Universität Kaiserslautern. 
His research interests include knowledge management, semantic desktop, electronic pen-input devices,
on-line and off-line handwriting recognition
and document analysis. He is a member of
the IAPR and frequent reviewer for international
journals, including IEEE Transactions
on Pattern Analysis and Machine Intelligence, IEEE Transactions on
Audio, Speech and Language Processing, Pattern Recognition, and
Pattern Recognition Letters.
\end{IEEEbiography}

\end{document}